\documentclass[a4paper,11pt]{article}

\usepackage{hyperref}
\usepackage{cite}
\usepackage{fourier}
\usepackage{color}
 \usepackage{graphicx}
\usepackage{url}
\usepackage[affil-it]{authblk}
\usepackage{amsmath}
\usepackage{wrapfig}

\usepackage[T1]{fontenc}
\usepackage{times}

\usepackage{rotating}
\usepackage{multirow}

\usepackage{diagbox}  

\usepackage{listings}
\usepackage{xcolor}
\definecolor{darkgreen}{rgb}{0.0, 0.5, 0.0}  

\lstdefinelanguage{json}{
    morestring=[b]",
    morestring=[b]',
    stringstyle=\color{black}, 
    keywordstyle=\color{blue},      
    commentstyle=\color{black},
    sensitive=true,
    morecomment=[s]{/*}{*/},
    morecomment=[l]{//},
    morekeywords={true,false,null},
    literate={:}{{\color{red}:}}1 
}

\lstdefinestyle{jsonstyle}{
    language=json,
    basicstyle=\footnotesize\ttfamily,
    keywordstyle=\color{blue},  
    commentstyle=\color{black},  
    stringstyle=\color{black},  
    numbers=left,
    numberstyle=\tiny\color{gray},
    stepnumber=1,
    numbersep=3pt,
    backgroundcolor=\color{white},
    showspaces=false,
    showstringspaces=false,
    showtabs=false,
    tabsize=2,
    morestring=[b]",
    morestring=[b]',
    literate={:}{{\color{black}:}}1
}

\begin{document}

\title{Benchmarking EfficientTAM on FMO datasets}

\author{Senem Aktas, Charles Markham,  John McDonald \&  Rozenn Dahyot}
\affil{Department of Computer Science Maynooth University Ireland}

\date{}
\maketitle
\thispagestyle{empty}

\begin{abstract}
Fast and tiny object tracking remains a challenge in computer vision and in this paper we first introduce a JSON metadata file associated with  four  open source datasets of Fast Moving Objects (FMOs) image sequences. In addition,   
we extend the description of the FMOs datasets with additional ground truth information in JSON format (called FMOX) with object size information. Finally we use our FMOX file to test a recently proposed foundational model for tracking (called EfficientTAM)  showing that its performance compares well with the pipelines originally  taylored for these FMO datasets. Our comparison of these state-of-the-art techniques on FMOX is provided with Trajectory Intersection of Union (TIoU) scores. 
The code and JSON is shared open source allowing FMOX to be accessible and usable for other machine learning pipelines aiming to process FMO datasets. 
\end{abstract}

\section{Introduction}

The tracking and detection of small and/or fast-moving objects (FMOs) remains relatively underexplored, particularly in comparison to research on general objects \cite{zhang2022tracking}. Specific challenges occur in such scenarios including motion blur, which can distort the appearance of objects and complicate the performance of general object detectors and trackers. Moreover, data annotation becomes more difficult due to the size and motion blur of FMOs, leading to a scarcity of available annotated datasets \cite{zhang2022tracking,yu20201st, rozumnyi2017world}. Despite these issues,
several public FMO datasets, including Falling Object \cite{kotera2020restoration}, FMOv2 \cite{rozumnyi2017world}, TbD \cite{kotera2019intra}, and TbD-3D \cite{rozumnyi2020sub} are now available (see Section~\ref{sec:SOTA:FMO:datasets}). 
Unlike larger objects, small objects often suffer from  reduced visibility and lower image cover rate, leading to fewer appearance cues and increased background interference. 
Considering both issues, fast and small object tracking, are therefore still challenging  for modern tracking techniques~\cite{zhang2022tracking,haalck2024tracking,yu20201st}.

In this work, we define and characterize the terms \textit{small} and \textit{fast} in the context of object tracking in videos.
Second, we provide a combined and extended dataset, which we term FMO eXtended (FMOX), combining the four datasets mentioned above with a more informative metadata encoding. This encoding is provided through a simplified JSON format, allowing the development of Machine Learning dataloaders and pipelines that operate over the entire unified dataset (Section~\ref{sec:FMOX}). Finally, we assess EfficientTAM (Efficient Track Anything Model) \cite{xiong2024efficient}, a recent state-of-the-art technique for tracking, on these datasets, benchmarking its performance against leading FMO specific techniques from the literature. 
Code and FMOX JSON are made available at \url{https://github.com/CVMLmu/FMOX/}.

\section{Fast-Moving Object Datasets}
\label{sec:SOTA:FMO:datasets}

The following  datasets\footnote{available at \url{https://cmp.felk.cvut.cz/fmo/}} are considered here to create FMOX.

\begin{figure}[h]
    \centering
    \begin{minipage}{0.49\textwidth}
        \centering
        \includegraphics[width=\textwidth]{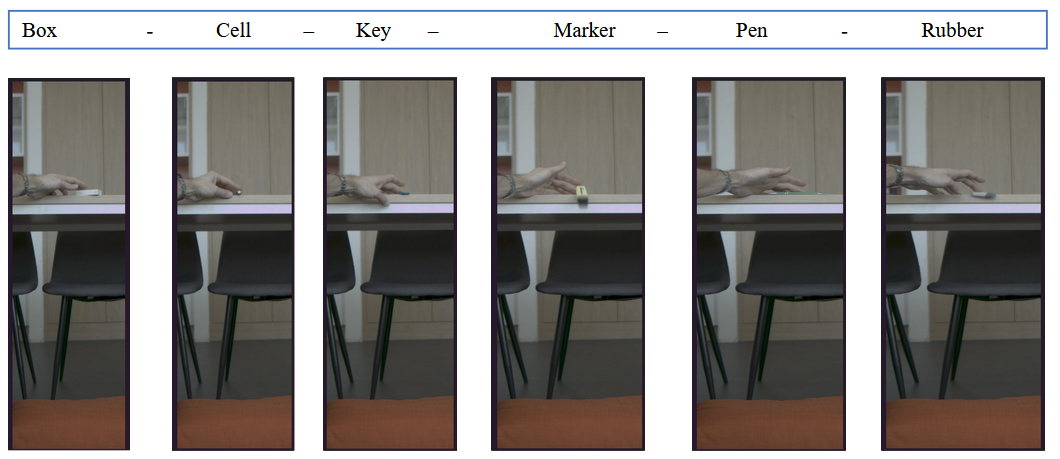}
        \caption{Falling Objects~\cite{kotera2020restoration} samples.}
        \label{fig:falling_db}
    \end{minipage}\hfill
    \begin{minipage}{0.49\textwidth}
        \centering
        \includegraphics[width=\textwidth]{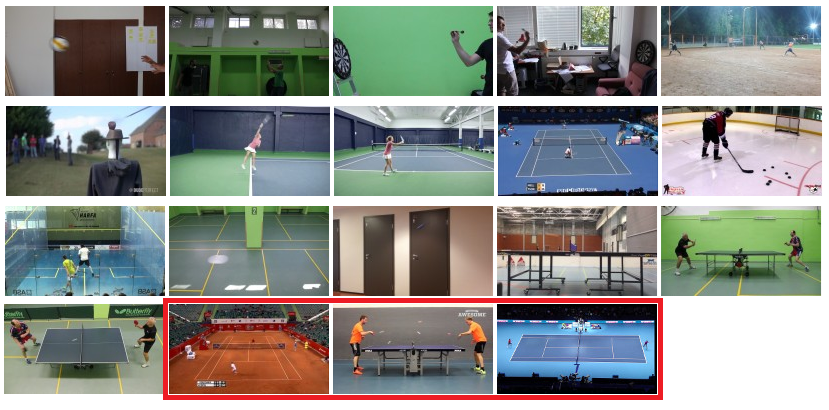}
        \caption{FMOv2~\cite{rozumnyi2017world} samples.}
        \label{fig:FMO_db}
    \end{minipage}
    \label{fig:two_images1}
\end{figure}

\paragraph{Falling Objects~\cite{kotera2020restoration}.} 
The dataset comprises static image frames from video sequences recorded at 25/30 Frame Per Second (FPS). It comprises a collection of 6 different objects, including box, cell, key, and rubber, which are dropped from a table, as shown in Figure \ref{fig:falling_db}. It includes two sets of images as high-speed and low-speed sequences, with ground truth provided in text format.

\paragraph{FMOv2~\cite{rozumnyi2017world}.} This dataset includes 19 sports sequences featuring a variety of small objects, primarily balls, such as volleyballs and tennis balls, but also objects such as darts and frisbees, with PNG images of varying lengths. The ground truths are provided in Matlab for each sequence, with object mask PNG format and its run length encoding compressed text file. In the dataset, FMO displacement is evenly spread between 0-150 pixels while bounding boxes between consecutive frames have nearly zero intersection each time. Figure \ref{fig:FMO_db} illustrates samples from the dataset, with the FMOv2 extension sequence samples highlighted in red. 

\paragraph{TbD~\cite{kotera2019intra}.}  This dataset comprises striking golf, tennis, volleyball and badminton balls as well as a falling cube and coin. 14 image sequences are provided with ground-truth trajectories including 12 sequences from  sports videos with mostly spherical objects and almost no appearance changes over time. These 30 FPS low-speed videos are  obtained from 240 FPS high-speed videos using temporal averaging. Semi-manual annotation are applied for creating ground truth by labeling the first frame  and applying  a tracker, and then correcting the annotations.

\begin{figure}[!h]
    \centering
    \begin{minipage}{0.48\textwidth}
        \centering
        \includegraphics[width=\textwidth]{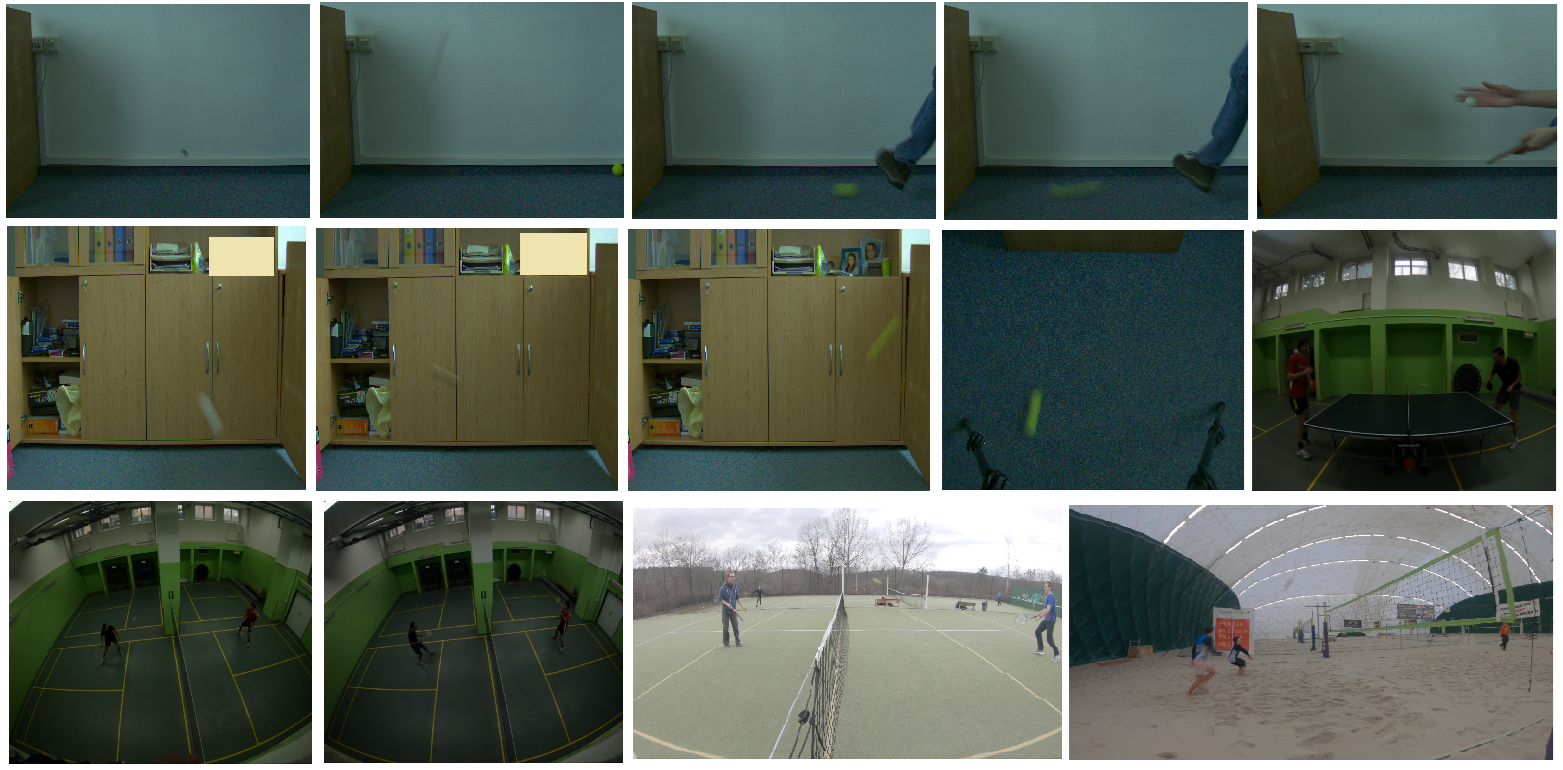}
        \caption{TbD~\cite{kotera2019intra} samples.}
        \label{fig:tbd_db}
    \end{minipage}\hfill
    \begin{minipage}{0.49\textwidth}
        \centering
        \includegraphics[width=\textwidth]{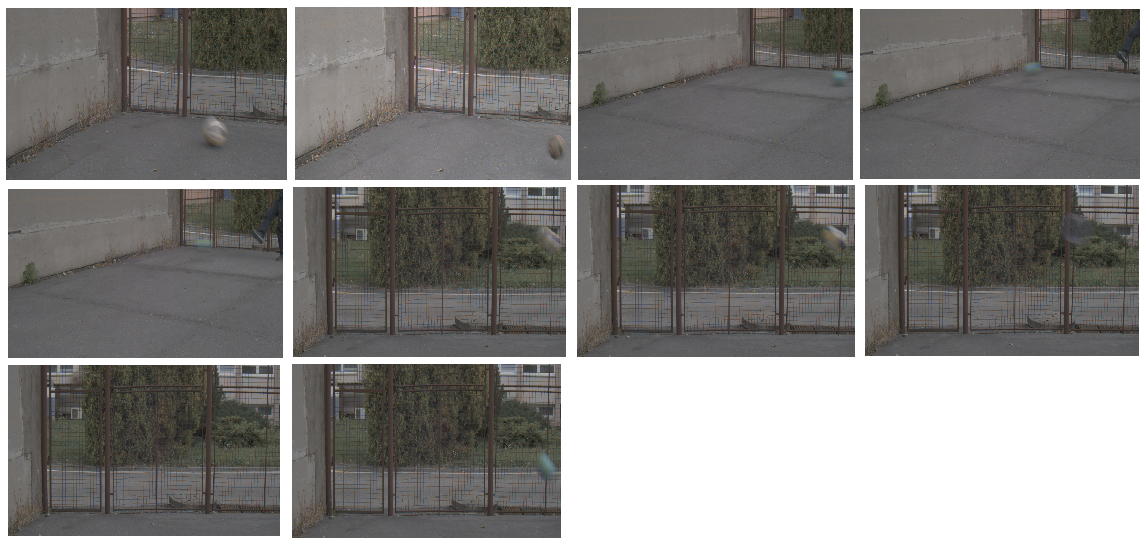}
        \caption{TbD-3D~\cite{rozumnyi2020sub} samples.}
        \label{fig:tbd3d_db}
    \end{minipage}
    \label{fig:two_images}
\end{figure}

\paragraph{TbD-3D~\cite{rozumnyi2020sub}.} Similar to TbD, the TbD-3D dataset includes 10 PNG image sequences, focusing on objects moving in 3D that undergo significant changes in appearance within a single low-speed videos. 9 of these sequences depict a spherical object, specifically a ball, while the last sequence features a circular object. 
The 6D poses (3D position and rotation) of FMOs are provided as ground truth with  manually annotated 3D object location (2D position and radius) and estimated 3D object rotation. 
Videos were captured in raw format with a high-speed camera operating at 240 FPS. Both high-speed and low-speed versions of the image sequences are provided. Ground truths are available in both MATLAB format and text format as trajectories for the low-speed image sequences. Figure \ref{fig:tbd3d_db} shows a sample frame from each sequence.

\section{FMOX}
\label{sec:FMOX}

In this section, we define FMOX, a JSON structured annotation format that simplifies the handling of the four public FMO datasets described in Section~\ref{sec:SOTA:FMO:datasets}). It is designed to be easily interpretable and compatible, enabling researchers and practitioners to quickly understand the structure and content of the data, allowing them to use the datasets with minimal effort. 
We also provide object size categories as additional labeling for the datasets. 
By considering size as a key factor in the annotations, we aim to enhance the relevance of the dataset for various research applications, such as small object detection and tracking.

\paragraph{Object Size.}\label{sec:object_size}
In the computer vision community, the term "small objects" characterizes as covering an area of $32 \times 32$ pixels or less \cite{tong2020recent, lin2014microsoft, zhang2022tracking}. For instance, the Microsoft COCO benchmark \cite{lin2014microsoft} categorizes objects into three size categories for evaluation as \texttt{small}, \texttt{medium}, and \texttt{large} based on the dimensions of the bounding boxes that encapsulate the objects. Object size categorization facilitates the analysis of detector and model performance across various object sizes \cite{zhu2016traffic}. 
Table \ref{tab:obj:size} presents object size categories as defined in the literature.
 Table \ref{tab:fmox_obj_size} presents our object size categories used in FMOX.

Table \ref{table:FMOX_Statistics} summarizes the FMOX statistics, primarily focusing on object size information. Additionally, the listing \ref{lst:fmox_structure} displays the structure of FMOX for the four analyzed FMO datasets.

\begin{table}[!h]
\begin{center}
\resizebox{\textwidth}{!}{ 
\begin{tabular}{|c|c|c|}
 \hline
 Study & Resolution & Categorization\\
 \hline
 Traffic sign \cite{zhu2016traffic} & $2048 \times 2048$ & \texttt{small}
$(0, 32]$, \texttt{medium} $(32, 96]$, \texttt{large} $(96, 400]$ \\

 \cite{tong2020recent} & & \texttt{small} $(0, 32]$, \texttt{medium} $(32, 96]$, \texttt{large} $(96, \infty]$\\
 
 TinyPerson \cite{yu20201st} & $1920 \times 1080$ & \texttt{tiny} $[2, 20]$ : \texttt{tiny1} $[2, 8]$, \texttt{tiny2} $[8,12]$, \texttt{tiny3} $[12, 20]$, \texttt{small} $[20, 32]$, \texttt{all} $[2,\infty]$ \\
 
 \cite{ying2025visible}  & & \texttt{extremely tiny} $[1 ,8 )$, \texttt{tiny} $[8 ,16 )$, \texttt{small} $[16,32)$ \\
  \hline
 \end{tabular}
 }
 \caption{Examples of object size category definitions using  side length of square bounding box.}\label{tab:obj:size}
 \end{center}
\end{table}

\begin{table}[!h]
\begin{center}
\begin{tabular}{|c|c|c|c|c|}
 \hline
 \texttt{Extremely Tiny} &   \texttt{Tiny}& \texttt{Small} & \texttt{Medium} & \texttt{Large}\\
 \hline
 \( [1 \times 1, 8 \times 8)\)& \( [8 \times 8,16 \times 16)\)& \( [16 \times 16,32 \times 32)\) & \( [32 \times 32, 96 \times 96)\) & \( [96 \times 96,\infty)\)\\
  \hline
 \end{tabular}
 \caption{FMOX object size categories.}\label{tab:fmox_obj_size}
 \end{center}
\end{table}

\paragraph{Fast Movement.} 

Fast-Moving Objects (FMOs) exhibit significant displacement that exceeds their size within the exposure time across consecutive frames \cite{rozumnyi2017world}. Movement of FMOs result in pronounced 6D pose changes in sequential frames; specifically, the 3D rotation of the object alters due to high angular velocity, causing it to appear partially visible and manifest as shadowy streaks \cite{rozumnyi2020sub,rozumnyi2017world}. According to \cite{rozumnyi2017world}, FMOs are perceived as translucent lines that are larger than their actual size, which is typically less than 100 pixels. Their motion in the $x, y, z$ planes leads to substantial changes in their 3D location, affecting their 2D position and radius, as well as the distance between the object and the camera \cite{rozumnyi2020sub}.

This streaking effect results in motion blur, complicating the ability to discern the object's features, such as its shape and edges, particularly in a single image \cite{rozumnyi2017world}. Evaluation results from their proposed model indicate that significant motion against a distinct background yields the best tracking outcomes. Furthermore, it has been noted \cite{rozumnyi2017world} that FMOs may go undetected when their motion is minimal or when the background color closely resembles that of the object. Additionally, local movements of large non-FMOs can sometimes be misidentified as FMOs.

Statistical analysis of FMO dataset \cite{rozumnyi2017world} demonstrated that in two consecutive frames the overlap of the ground truth bounding boxes is zero. The distance between the center of the object is on average ten times higher than non-FMO datasets and the displacement is uniformly spread between 0-150 pixels. However,  in two consecutive frames of non-FMO datasets, ground truth bounding boxes overlap is close to one or above 0.5 in 94\% of cases, and the displacement is below 10 pixels in 91\% of cases. In small and fast-moving object benchmark \cite{zhang2022tracking}, it is expected that target center moves by at least 50\% of its size. 

\paragraph{FMOX Structure.}

The four examined FMO datasets exhibit different formats, including variations in annotation styles such as text annotations, segmentation masks and Matlab versions. These variations in annotations could limit the reproducibility of experiments, which inspired us to develop an easy-to-use JSON format annotation called FMOX, with the goal of improving accessibility, readability, and usability for users. The JSON structure (shown  in Listing \ref{lst:fmox_structure}) contains objects' bounding boxes as $(x_1,y_1,x_2,y_2)$ format and object size category. To obtain the bounding boxes  for FMOv2, the object mask images provided are processed with OpenCV's contour detection function \texttt{findContours} (with parameters \texttt{RETR\_EXTERNAL}, \texttt{CHAIN\_APPROX\_SIMPLE} and threshold value of 70). For the TbD dataset, ground-truth trajectory text annotations are utilized to obtain bounding boxes which consist of object annotations for the entire sequence of frames (and not only for FMO frames). 
For the Falling Object and TbD-3D datasets, the data loading component of the repository \texttt{fmo-deblurring-benchmark}\footnote{\url{https://github.com/rozumden/fmo-deblurring-benchmark} \cite{rozumnyi2021defmo}}  is leveraged.

\begin{lstlisting}[style=jsonstyle, caption={Structure of FMOX},label={lst:fmox_structure}]
{ "databases": [ {
      "dataset_name": "Falling_Object",
      "version": "1.0",
      "description": "Falling_Object annotations.",
      "sub_datasets": [
                {"subdb_name": "v_box_GTgamma",
                 "images": [
                    {
                        "img_index": 1,
                        "image_file_name": "00000027.png",
                        "annotations": [
                            {
                                "bbox_xyxy": [161, 259, 245, 333],
                                "object_wh": [84, 74],
                                "size_category": "medium"
                            }
                        ]},
                    {
                        "img_index": 2,
                        "image_file_name": "00000028.png",
                        "annotations": [ ---- ]
                    }   ] } -------- ]  }, ------- }
\end{lstlisting}

\begin{table}[h!]
\centering
\resizebox{\textwidth}{!}{ 
\begin{tabular}{|l|c|c|c|c|c|}
\hline
\multicolumn{2}{|c|}{\multirow{4}{*}{Sequence Name}} & \multicolumn{4}{c|}{Analysis} \\
\cline{3-6}
\multicolumn{1}{|c}{} & & \multirow{4}{8ex}{Total Frame Number} & \multirow{4}{4em}{FMO Exists Frame Number} & \multirow{4}{4em}{\textbf{(Ours)} TIoU $(\uparrow)$} & \multirow{4}{4em}{Object Size Levels} \\ [9ex]
\hline
\multirow{8}{*}{\begin{turn}{90}Falling Object \end{turn}}  
    &  &   &   &  &  \\ 
    & v\_box\_GTgamma & 62 & 22 &  \textbf{0.904} & \{'medium': 22\} \\
    & v\_cell\_GTgamma & 62 & 14 &  \textbf{0.730} & \{'medium': 14\} \\
    & v\_key\_GTgamma & 62 & 19 &  \textbf{0.651} & \{'medium': 19\} \\
    & v\_marker\_GTgamma & 62 & 11 &  \textbf{0.799} & \{'medium': 11\} \\
    & v\_pen\_GTgamma & 62 & 13 &  \textbf{0.558} & \{'large': 8, 'medium': 5\} \\
    & v\_rubber\_GTgamma & 62 & 15 &  \textbf{0.614} & \{'medium': 15\} \\
    &  &   &   &  &  \\ 
\hline
\multirow{19}{*}{\begin{turn}{90}FMOv2 \end{turn}}
     &  &   &   &  &  \\ 
     & atp\_serves$^+$ & 655 & 463 &  0.135 * & \{'extremely\_tiny': 1, 'small': 79, 'tiny': 383\} \\
     & blue & 53 & 21 & \textbf{0.775} & \{'large': 1, 'medium': 20\} \\
     & darts1 & 75 & 51 &  \textbf{0.738} & \{'large': 3, 'medium': 33\} \\
     & darts\_window1 & 50 & 9 &  0.023 * & \{'medium': 5\} \\
     & frisbee$^+$ & 100 & 68 &  0.490 & \{'large': 16, 'medium': 4\} \\
     & hockey$^+$ & 350 & 323 &  \textbf{0.527} & \{'extremely\_tiny': 48, 'small': 6, 'tiny': 7\} \\
     & more\_balls $^+$ & 300 & 287 &  \textit{Not applied} & \{'medium': 129, 'small': 1112, 'tiny': 49\} \\
     & ping\_pong\_paint & 120 & 111 &  0.036 & \{'extremely\_tiny': 1, 'medium': 68, 'small': 6, 'tiny': 1\} \\
     & ping\_pong\_side$^+$ & 445 & 444 & \textbf{0.629} & \{'extremely\_tiny': 2, 'medium': 172, 'small': 183, 'tiny': 79\} \\
     & ping\_pong\_top$^+$ & 350 & 350 & 0.396 & \{'extremely\_tiny': 1, 'large': 2, 'medium': 242, 'small': 59, 'tiny': 148\} \\
     & softball & 96 & 35 & 0.009 * & \{'medium': 14, 'small': 13, 'tiny': 1\} \\
     & squash & 250 & 242 & \textit{NAN} & \{'extremely\_tiny': 129, 'tiny': 5\} \\
     & tennis1 & 116 & 91 &  \textit{NAN} & \{'extremely\_tiny': 63, 'tiny': 1\} \\
     & tennis2 & 278 & 274 & 0.005 & \{'extremely\_tiny': 151, 'small': 6, 'tiny': 62\} \\
     & tennis\_serve\_back$^+$ & 156 & 78 & 0.312 & \{'extremely\_tiny': 31, 'small': 10, 'tiny': 18\} \\
     & tennis\_serve\_side & 68 & 35 & \textbf{0.821} & \{'medium': 1, 'small': 12, 'tiny': 5\} \\
     & volleyball1 & 50 & 33 &  \textbf{0.905} & \{'large': 12, 'medium': 1\} \\
     & volleyball\_passing & 66 & 66 & \textbf{0.895} & \{'large': 4, 'medium': 62\} \\
     & william\_tell & 119 & 67 & \textbf{0.783} & \{'extremely\_tiny': 1, 'large': 7, 'medium': 7, 'small': 5, 'tiny': 12\} \\
    &  &   &   &  &  \\ 
\hline
\multirow{12}{*}{\begin{turn}{90}TbD \end{turn}}
    &  &   &   &  &  \\ 
    & VS\_badminton\_white\_GX010058-8 & 125 & 40 & 0.010 * & \{'tiny': 56, 'extremely\_tiny': 36, 'small': 27, 'medium': 6\} \\ 
    & VS\_badminton\_yellow\_GX010060-8 & 125 & 57 & 0.265 & \{'tiny': 63, 'extremely\_tiny': 36, 'small': 19, 'medium': 7\} \\
    & fall\_cube & 28 & 20 & \textbf{0.902} & \{'medium': 4, 'small': 2, 'tiny': 5, 'extremely\_tiny': 17\} \\
    & hit\_tennis & 57 & 30 & \textbf{0.878} & \{'extremely\_tiny': 47, 'tiny': 9, 'small': 1\} \\
    & hit\_tennis2 & 26 & 26 & 0.094 * & \{'extremely\_tiny': 4, 'tiny': 14, 'small': 5, 'medium': 3\} \\
    & VS\_pingpong\_GX010051-8 & 95 & 58 & \textbf{0.756} & \{'small': 36, 'tiny': 46, 'extremely\_tiny': 13\} \\
    & VS\_roll\_golf-gc-12 & 16 & 16 & \textbf{0.858} & \{'small': 3, 'medium': 5, 'tiny': 3, 'extremely\_tiny': 5\} \\
    & VS\_tennis\_GX010073-8 & 118 & 38 & \textbf{0.807} & \{'small': 66, 'tiny': 32, 'extremely\_tiny': 20\} \\
    & throw\_floor & 73 & 40 &  0.003 * & \{'medium': 15, 'large': 1, 'small': 6, 'tiny': 7, 'extremely\_tiny': 44\} \\
    & throw\_soft & 75 & 60 &  0.008 * & \{'small': 16, 'large': 1, 'medium': 13, 'tiny': 7, 'extremely\_tiny': 38\} \\
    & throw\_tennis & 71 & 45 &  0.003 * & \{'medium': 16, 'small': 19, 'tiny': 9, 'extremely\_tiny': 27\} \\
    & VS\_volleyball\_GX010068-12 & 72 & 41 & \textbf{0.872} & \{'small': 11, 'medium': 5, 'tiny': 25, 'extremely\_tiny': 31\} \\
    &  &   &   &  &  \\ 
\hline
\multirow{10}{*}{\begin{turn}{90}TbD-3D \end{turn}}
    &  &   &   &  &  \\ 
    & HighFPS\_GT\_depth2 & 48 & 48 & \textbf{0.860} & \{'large': 2, 'medium': 46\} \\
    & HighFPS\_GT\_depthb2 & 81 & 81 &  \textbf{0.823} & \{'medium': 81\} \\
    & HighFPS\_GT\_depthf1 & 46 & 46 & \textbf{0.833} & \{'medium': 46\} \\
    & HighFPS\_GT\_depthf2 & 50 & 50 & \textbf{0.816} & \{'medium': 50\} \\
    & HighFPS\_GT\_depthf3 & 37 & 37 &  \textbf{0.816} & \{'medium': 37\} \\
    & HighFPS\_GT\_out1 & 57 & 57 & \textbf{0.899} & \{'large': 1, 'medium': 56\} \\
    & HighFPS\_GT\_out2 & 50 & 50 &  \textbf{0.909} & \{'medium': 50\} \\
    & HighFPS\_GT\_outa1 & 47 & 47 & \textbf{0.923} & \{'large': 14, 'medium': 33\} \\
    & HighFPS\_GT\_outb1 & 41 & 41 & \textbf{0.830} & \{'medium': 41\} \\
    & HighFPS\_GT\_outf1 & 60 & 60 & \textbf{0.895}  & \{'medium': 60\} \\
    &  &   &   &  &  \\ 
\hline

\end{tabular}
}
\caption{FMOX dataset information with Trajectory-Intersection of Union (TIoU)  results computed  on each sequence with EfficientTAM \cite{xiong2024efficient} (column labelled \textbf{(Ours)}; TIoU above 0.5 are in bold font). \textit{Not applied}: multi object sequence (request multi initialization with no id for comparison); *: tracker could not initialized due to high motion blur. \textit{NAN} is one of the outputs of the TIoU function used. Sequences noted $^+$ indicate that multiple objects occur but labels  provided are not with object instance ids.}
\label{table:FMOX_Statistics}
\end{table}

\section{EfficientTAM performance on FMOX}

To provide a baseline measure of performance, we test Efficient Track Anything Model (EfficientTAM) \cite{xiong2024efficient}  on FMOX.
No additional training is conducted to the pretrained model \texttt{efficienttam\_s}   which we use with its default parameters\footnote{\url{https://github.com/yformer/EfficientTAM}.}. 
The FMOX ground-truth annotation  from the first image of each sequence is used  to initialize the EfficientTAM with a target to track in that sequence. Both point (chosen as the center of the bounding box) and bounding box initializations were assessed for initializing the target.
EfficientTAM  first performs  segmentation on the first frame to then track the detected region in following frames.  When the initialization is point-based, it is often observed that the detection of  the target object fails whereas when using the bounding box, the segmentation correctly captures the object of interest. Hence, tracking is tested with bounding box initialization on the first frame.
Note however, that even with the bounding box, if the frame exhibits strong motion blur, EfficientTAM can still fail  to segment the object to be tracked on the first frame.
The performance of EfficientTAM, as measured by the Trajectory-Intersection of Union (TIoU) \cite{rozumnyi2021IJCV}, is reported in Tables  \ref{table:FMOX_Statistics} and \ref{tab:EfficientTAM_performance}, as TIoU is employed in studies of fast-moving objects to evaluate FMO datasets.

To evaluate the performance of EfficientTAM on four datasets, we utilized the 
Defmo\footnote{\url{https://github.com/rozumden/DeFMO}.} pipeline to perform TIoU \cite{rozumnyi2021IJCV} calculations with FMOX. 
To feed the pipeline, FMOX object bounding boxes are transformed into trajectories by calculating the center coordinates of the bounding boxes 
and then interpolating according to the number of segment (nsplits) parameter.
 Coding details are available on GitHub \url{https://github.com/CVMLmu/FMOX/}.

\begin{table}[!h]
\begin{center}
\resizebox{\textwidth}{!}{ 
\begin{tabular}{|c|c|c|c|c|c|}
 \hline
 \diagbox{Datasets}{Studies}  & \multirow{1}{12ex}{Defmo \cite{rozumnyi2021defmo}} & \multirow{1}{12ex}{FmoDetect \cite{rozumnyi2021fmodetect}} & \multirow{1}{12ex}{TbD \cite{kotera2019intra}} & \multirow{1}{12ex}{TbD-3D  \cite{rozumnyi2020sub}} & \multirow{1}{12ex}{\textbf{(Ours)} with EfficientTAM}\\  [5ex]
 \hline
 Falling Object & $0.684^{**}$ &  N/A &  0.539 & 0.539 & $0.7093^{*}$ \\  \hline
 
 TbD &  $0.550^{**}$ &  (a) 0.519  (b) $0.715^{*}$ &  0.542 & 0.542 & 0.4546\\   \hline
  TbD-3D &  $0.879^{*}$ &  N/A & 0.598 & 0.598 & $0.8604^{**}$\\
   \hline
 \end{tabular}
 } \\
 (a) real-time with trajectories estimated by the network, (b) with the proposed deblurring, 
 (N/A):   not defined \\
 \caption{Average TIoU $(\uparrow)$ performance comparison of our results with EfficientTAM \cite{xiong2024efficient}  on FMO datasets with FMOX (column \textbf{(Ours)}). For each dataset, $^*$ indicate best result and $^{**}$ second best result. Other Average TIoUs shown are directly extracted from the cited papers.}
 \label{tab:EfficientTAM_performance}
 \end{center}
\end{table}

\begin{figure}[!h]
\begin{center}
 \includegraphics[width=0.9\textwidth, height=0.6\textwidth]{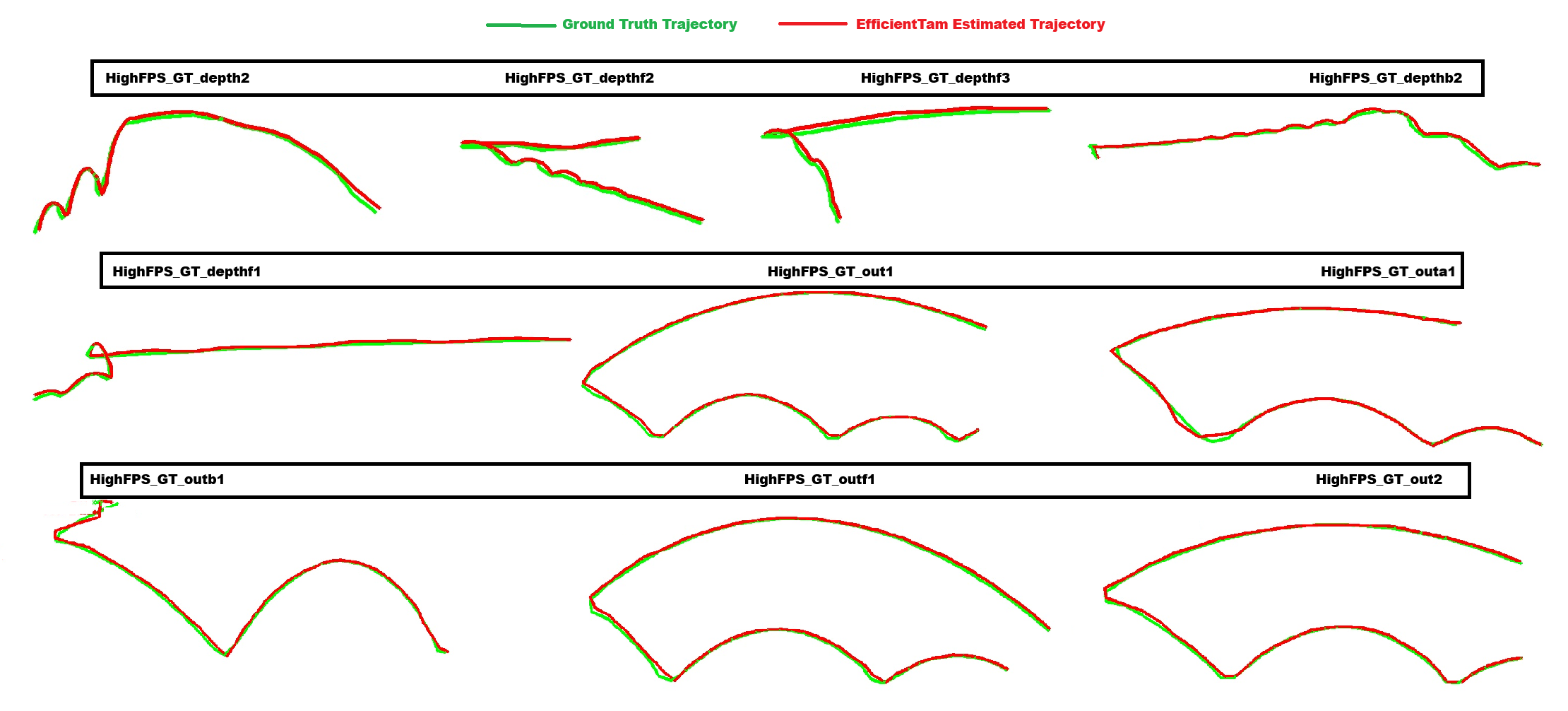}
 \end{center}
 \caption{EfficientTAM  estimated trajectories on TbD-3D dataset. Green color indicates ground truth trajectory while red color for EfficientTAM estimated trajectory. TIoU values are above 0.81 for all sequences. Objects (mostly balls) quite big, while having motion blur still object is pretty visible all along sequences.}
   \label{fig:tbd3d_db_traj}
\end{figure}

Overall  EfficientTAM performs very well in particular with the Falling Object  and TbD-3D  datasets (cf. Fig. \ref{fig:tbd3d_db_traj} and Tab. \ref{tab:EfficientTAM_performance}).
However, a general issue with the FMOv2 and TbD datasets is that some sequences, could not be initialized due to strong motion-blurred FMOs (see notation asterisk (*) in Table \ref{table:FMOX_Statistics}). 
We believe that initializing the EfficientTAM  with less motion-blurred FMOs frames would yield higher scores. Moreover, for the FMOv2 dataset, some sequences contain multiple FMOs, such as \texttt{frisbee} and \texttt{more\_balls}. However, without unique IDs for these objects, it is not possible to effectively compare the ground truth with the estimated results. In the \texttt{frisbee} sequence, two frisbee objects travel nearly the same distance and direction but at different times. We initialized the tracker for the first object, which covers almost half of the trajectory, resulting in a TIoU value of approximately 0.5. In the \texttt{more\_balls} sequence, multiple balls appear and disappear repeatedly, which is why we have not included this sequence in our evaluation for the time being. 
Also, in FMOX, we have corrected several masks  in the \texttt{ping\_pong\_paint} sequence, which contained only a small mask of ball from a different tennis game that was interfering with the tracking initialization. Similarly, the \texttt{william\_tell} sequence was distorted by extra masks --- specifically, traces left by pieces of the apple that was shot --- misdirecting the trajectory of the main target (bullet). After making these mask corrections, the TIoU improved to 0.783.

\section{Conclusion}

We propose an enhanced metadata description file, FMOX, associated with four video datasets featuring Fast Moving Objects (FMOs). Using FMOX, we evaluate the recently released foundation model, EfficientTAM, for tracking FMOs. Our experimental results demonstrate its competitive performance and limitations, including difficulties in initializing tracking for strongly motion-blurred objects, in challenging scenarios. Notably, EfficientTAM achieves superior performance without requiring specialized training or modifications to its default parameters, yielding average Trajectory-Intersection over Union (TIoU) scores of 0.7093 for Falling Objects and 0.8604 for TbD-3D datasets.
These results showcase its effectiveness in tracking FMOs. Future work will investigate additional metrics to TIoU for performance assessment of trackers with FMOX, as well as evaluating  impact of  object size and motion. Some recommendations in using EfficientTAM can be made from this experiments such as best performance is obtained when the tracker is initialized on a non-blurry object in the image. Moreover in the cases where the tested sequence displays several instances of the same object (e.g. tennis balls),  then the tracker can be distracted by a competing instance leading to a low TIoU for the sequence.  

\section*{Acknowledgments}

This research was supported by funding through the  Maynooth University Hume Doctoral Awards. For the purpose of Open Access, the author has applied a CC BY public copyright licence to any Author Accepted Manuscript version arising from this submission.


\end{document}